\documentclass[10pt, twocolumn]{amsart}
\usepackage{multicol}
\usepackage{blindtext}
\usepackage{graphicx}
\usepackage{amssymb}
\usepackage[T1]{fontenc}

\usepackage{mathptmx}
\usepackage{layout}
\usepackage{marvosym}
\usepackage{enumitem}
\usepackage{tikz}
\usetikzlibrary{arrows,backgrounds}

\usepackage{xcolor}

\usepackage{titlesec}

\usepackage[hidelinks]{hyperref}

\usepackage[final]{pdfcomment}
\usepackage{comment}

\makeatletter
\def\specialsection{\@startsection{section}{1}%
  \z@{\linespacing\@plus\linespacing}{.5\linespacing}%
  {\normalfont}}
\def\section{\@startsection{section}{1}%
  \z@{.7\linespacing\@plus\linespacing}{.5\linespacing}%
  {\normalfont\scshape}}
\makeatother

\newcommand{\SEC}[1]{
  \titlelabel{\textbf{\thesection}\quad}
  \section{\textbf{#1}}
  \vspace{-3mm}
}
\newcommand{\SUBSEC}[1]{
  \titlelabel{\textbf{\thesubsection}\quad}
  \subsection{\textbf{#1}}~\\\vspace{-5mm}}

\newcommand{\Dir}{\mbox{Dirichlet}}
\newcommand{\Mult}{\mbox{Multinomial}}

\newcommand{\LL}{\mathcal{L}}
\newcommand{\E}{\mathbb{E}}

\begin{document}
\title{}
\twocolumn[
\begin{center}
  {
    \huge
    \textbf{Algorithms of the LDA model}
  }
  \\
  \vspace{0.5cm}
  {
    \large
    \textit{Jaka \v{S}peh, Andrej Muhi\v{c}, Jan Rupnik}
    \\
    Artificial Intelligence Laboratory
    \\
    Jo\v{z}ef Stefan Institute
    \\
    Jamova cesta 39, 1000 Ljubljana, Slovenia
    \\
    e-mail: \{jaka.speh, andrej.muhic, jan.rupnik\}@ijs.si
  }
\end{center}
\vspace{0.7cm}]


  \begin{center}
    \textbf{ABSTRACT}
  \end{center}

{\bfseries
    We review three algorithms for Latent Dirichlet Allocation
    (LDA). Two of them are variational inference algorithms: Variational
    Bayesian inference and Online Variational Bayesian inference and one is
    Markov  Chain Monte Carlo (MCMC) algorithm -- Collapsed  Gibbs
    sampling.
    We compare their time complexity and performance. We find
    that online variational Bay\-esian inference is the fastest algorithm and
    still returns reasonably good results. 
}

  \vspace{-3mm}

  \SEC{INTRODUCTION}

  Nowadays big corpora are used daily. People often
  search through huge numbers of documents either in libraries or online, using
  web search engines. Therefore, we need efficient algorithms 
  that enable us efficient information retrieval.

  Sometimes appropriate documents are hard to find, especially if you do not have
  the exact title. One solution is to search using keywords. As many documents do not have keywords, 
  we need to know what certain document is talking about.
  Therefore, we would like to tag documents with appropriate keywords by clustering them according to their
  topics. 

  As the size of corpus increases, manual annotation is not an option. We would
  like that computers process documents and find their topics automatically. That can be done
using  machine learning.

  Probabilistic graphical models such as Latent Dirichlet Allocation (LDA)
  allow us to describe a document in
  terms of probabilistic distributions over
  topics, and these topics in terms of distributions over words. In order to
  obtain documents topics and corpus topics (distributions over words), we need to
  compute posterior distribution. Unfortunately, the posterior is intractable to
  compute and one must appeal to approximate posterior inference.

  Modern approximate posterior inference algorithms fall into two
  categories: sampling approaches and optimization approaches. Sampling
  approaches are usually based on MCMC
  sampling. Conceptual idea of the methods is to generate independent samples
  from posterior
  and then reason  about documents and corpus topics. Whereas optimization
  approaches are usually based on variational inference, also called Variational
  Bayes (VB) for Bayesian models. Variational Bayes methods optimize
  closeness, in Kullback-Leibler divergence, of simplified parametric distribution
  to the posterior.

  In this paper, we compare one MCMC and two VB algorithms for approximating
  posterior distribution. In the subsequent sections, we formally introduce LDA
  model and algorithms. We study performance of algorithms and make comparisons
  between them. For training and testing set we use articles from Wikipedia.
  We show that Online Variational Bayesian inference is the fastest
  algorithm. However the accuracy is lower than in the other two,
  but the results are still good enough for practical use.

  \SEC{LDA MODEL}

  Latent Dirichlet Allocation \cite{Blei:2003} is a Bayesian probabilistic
  gra\-phical model,
  which is regularly used in topic modeling.
  It assumes $M$ documents are build
  in a following fashion. First, a collection of $K$ topics (distributions over
  words) are drawn from a Dirichlet distribution,
  $\varphi_k \sim \Dir(\beta)$. Then for $m$-th document, we:
  \begin{figure}[!ht]
    \centering
      \begin{tikzpicture}[scale=0.7]

        \tikzstyle{post}=[->,shorten >=0pt,>=stealth',thick]
        \foreach \x/\y in {-2.2/1.5, 1.6/1.5, 4/1.5, 6.4/5.1,
          3/5.1} {
          \draw[thick] (\x, \y) circle (0.8);
        }

        \draw[thick, fill=gray] (6.4,1.5) circle (0.8);

        \draw[very thick] (0, -0.8) rectangle (8, 3.2);
        \draw[very thick] (2.9, 0) rectangle (7.5, 2.8);
        \draw[very thick] (5, 3.6) rectangle (8, 6.4);

        \node at (1.6, 1.5) {$\theta$};
        \node at (4, 1.5) {$z$};
        \node at (6.4, 1.5) {$w$};
        \node at (-2.2, 1.5) {$\alpha$};
        \node at (6.4, 5.1) {$\varphi$};
        \node at (3, 5.1) {$\beta$};

        \node at (7.7, -0.4) {$M$};
        \node at (7.2, 0.4) {$N$};
        \node at (7.7, 4) {$K$};

        \foreach \x/\y in {-1.4/0.8, 2.4/3.2, 4.8/5.6} {
          \draw[->, post] (\x, 1.5) -- (\y, 1.5);
        }

        \draw[->, post] (3.8, 5.1) -- (5.6, 5.1);
        \draw[->, post] (6.4, 4.3) -- (6.4, 2.3);
      \end{tikzpicture}
    \caption{Plate notation of LDA.}
    \label{fig:plate}
  \end{figure}
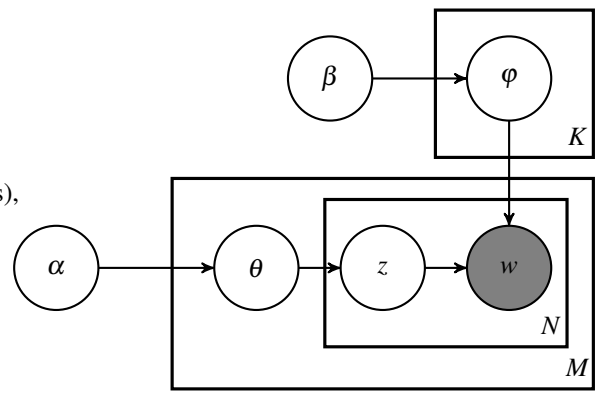
  
  \vspace{3cm}
  \begin{enumerate}[label={\arabic*.}]
  \item Choose a topic distribution $\theta_m \sim \Dir(\alpha)$.
  \item For each word $w_{m, n}$ in $m$-th document:
    \begin{enumerate}[label={\roman*.}]
    \item choose a topic of the word \goodbreak
      $z_{m, n} \sim \Mult(\theta_m)$,
    \item choose a word $w_{m, n} \sim \Mult(\varphi_{z_{m, n}})$.
    \end{enumerate}
  \end{enumerate}
  LDA can be graphically presented using plate notation (Figure
  \ref{fig:plate}).

  Probability of the LDA model is
  \begin{align*}
    &p(\mathbf{w}, \mathbf{z}, \mathbf{\theta}, \mathbf{\varphi}
    \mid \alpha,\beta) =
    \\
    &\prod_{k = 1}^K \hspace{-0.5mm}
    p(\varphi_k | \beta)
    \prod_{m = 1}^M \hspace{-1mm}
    \left(\hspace{-1mm}
      p(\theta_m | \alpha)
      \prod_{n = 1}^{N_m}
      p(z_{m, n} | \theta_m) p(w_{m, n} | \varphi_{z_{m, n}})
    \right).
  \end{align*}

  We can analyse a corpus of documents by computing the posterior distribution
  of the hidden variables ($\mathbf{z}, \mathbf{\theta}, \mathbf{\varphi}$)
  given a document ($\mathbf{w}$). This posterior reveals latent structure in \pdfcomment{$\mathbf{w}$) brez oklepajev?}
  the corpus that can be used for prediction or data exploration. Unfortunately,
  this distribution cannot be computed directly \cite{Blei:2003}, and is usually
  approximated using Mar\-kov Chain Monte Carlo
  (MCMC) methods or variational inference.

  \SEC{ALGORITHMS}

  In the following subsections, we will derive one MCMC algorithm and two
  variational Bayes algorithms for the approximation of the posterior
  inference.

  \SUBSEC{Collapsed Gibbs sampling}

  In the Collapsed Gibbs sampling we first integrate $\mathbf{\theta}$ and
  $\mathbf{\varphi}$ out.
  \[
  p(\mathbf{z}, \mathbf{w} \mid \alpha, \beta)
  =
  \int_{\mathbf{\theta}} \int_{\mathbf{\varphi}}
  p(\mathbf{z}, \mathbf{w}, \mathbf{\theta}, \mathbf{\varphi}
  \mid \alpha, \beta)\;
  d\hspace{-0.2mm}\mathbf{\theta}\,
  d\hspace{-0.2mm}\mathbf{\varphi}.
  \]
  The goal of Collapsed Gibbs sampling here is to approximate the distribution
  $p(\mathbf{z} \mid \mathbf{w}, \alpha, \beta)$. Conditional probability
  $p(\mathbf{w} \mid \alpha, \beta)$ does not depend on $\mathbf{z},$ therefore
  Gibbs sampling equations can be derived from
  $p(\mathbf{z}, \mathbf{w} \mid \alpha, \beta)$ directly. Specifically, we are
  interested in the following conditional probability
  \begin{align*}
    p(z_{m, n} \mid \mathbf{z}_{\lnot (m, n)}, \mathbf{w}, \alpha, \beta),
  \end{align*}
  where $\mathbf{z}_{\lnot (m, n)}$ denotes all $z$-s but $z_{m, n}$. Note
  that for Collapsed Gibbs sampling we need only to sample a value for $z_{m, n}$
  according to the above probability. Thus we only need the\pdfcomment{value of izbrisano}
  probability mass function up to scalar multiplication. So, the distribution can be simplified \cite[page 22]{Gregor:2005} as:
  \begin{align}
    \label{eq:1}
    &p(z_{m, n} = k \mid \mathbf{z}_{\lnot (m, n)}, \mathbf{w}, \alpha, \beta)
    \propto
    \\
    \nonumber
    &\frac{n_{k, \lnot (m, n)}^{(v)} + \beta}
    {\sum_{v = 1}^V (n_{k, \lnot (m, n)}^{(v)} + \beta)}~
    (n_{m, \lnot (m, n)}^{(k)} + \alpha),
  \end{align}
  where $n_k^{(v)}$ refers to the number of times that term $v$ has been
  observed with topic $k$, $n_m^{(k)}$ refers to the number of times that topic
  $k$ has been observed with a word of document $m$, and
  $n_{\;\cdot\;, \lnot(m, k)}^{(\;\cdot\;)}$ indicate that the $n$-th token in \pdfcomment{., mogoče bolje eksplicitno}
  $m$-th document is excluded from the corresponding $n_k^{(v)}$ or
  $n_m^{(k)}$.

  Corpus and document topics can be obtained by \cite[page 23]{Gregor:2005}:
  \[
    \varphi_{k, v}
    =
    \frac{n_k^{(t)} + \beta}{\sum_{v = 1}^V (n_k^{(t)} + \beta)},
\quad
    \theta_{m, k}
    =
    \frac{n_{m}^{(k)} + \alpha}{\sum_{k = 1}^K (n_{m}^{(k)} + \alpha)}.
  \]

  In Collapsed Gibbs sampling algorithm, we need to remember values of three
  variables: $z_{m, n}$, $n_m^{(k)},$ and $n_k^{(v)},$ and some sums of these
  variables for efficiency. The algorithm first initializes $\mathbf{z}$ and
  computes $n_m^{(k)}$, $n_k^{(v)}$ according to the initialized values. Then in
  one iteration of the algorithm we go over all words of all documents, sample
  values of $z_{m, n}$ according to Equation \eqref{eq:1}, and recompute $n_m^{(k)}$,
  $n_k^{(v)}$. Then one has to decide when (from which iteration/s) to take a
  sample or samples 
  and which criteria to choose to check if Markov chain has
  converged.

  \SUBSEC{Variational Bayesian inference}

  This algorithm was proposed in the original LDA paper~\cite{Blei:2003}.

  In Variational Bayesian inference (VB) the true posterior is approximated by a
  simpler distribution $q(\mathbf{z}, \mathbf{\theta}, \mathbf{\phi})$, which is
  indexed by a set of free parameters \cite{Jordan:1999}. We choose a fully
  factorized distribution $q$ of the form
  \begin{align*}
    q(z_{m, n} = k)
    &= \psi_{m, n, k},
    \\
    q(\theta_m)
    &= \Dir(\theta_m \mid \gamma_m),
    \\
    q(\varphi_k)
    &= \Dir(\varphi_k \mid \lambda_k).
  \end{align*}
  The posterior is parameterized by $\mathbf{\psi}$, $\mathbf{\gamma}$ and
  $\mathbf{\lambda}$. We refer to $\mathbf{\lambda}$ as corpus topics and
  $\mathbf{\gamma}$ as documents topics.

  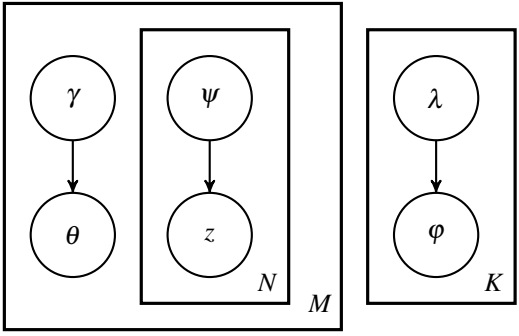
\begin{figure}[!ht]
    \centering
    \begin{tikzpicture}[scale=0.7]
      \tikzstyle{post}=[->,shorten >=0pt,>=stealth',thick]
      \draw[very thick] (0, 0) rectangle (2.8, -5.2);
      \draw[very thick] (-2.6, 0.5) rectangle (3.8, -5.7);
      \draw[very thick] (4.3, 0) rectangle (7.1 , -5.2);

      \foreach \x/\y/\name in {1.3/-1.3/$\psi$, 1.3/-3.9/$z$, -1.3/-1.3/$\gamma$,
        -1.3/-3.9/$\theta$, 5.6/-1.3/$\lambda$, 5.6/-3.9/$\varphi$
      }{
        \draw[thick] (\x, \y) circle (0.8);
        \node at (\x, \y) {\name};
      }

      \node at (2.4, -4.8) {$N$};
      \node at (6.7, -4.8) {$K$};
      \node at (3.4, -5.2) {$M$};

      \foreach \x/\y/\z/\u in {1.3/-2.1/1.3/-3.1,
        -1.3/-2.1/-1.3/-3.1,
        5.6/-2.1/5.6/-3.1
      }
      {
        \draw[->, post] (\x, \y) -- (\z, \u);
      }
    \end{tikzpicture}
    \caption{Plate notation of parameterized distribution $q$.}
    \label{fig:two}
  \end{figure}

  The parameters are optimized to maximize the Evidence Lower Bound (ELBO):
  \begin{align}
    &\log p(\mathbf{w} \mid \alpha, \beta)
    \geq
    \LL(\mathbf{w}, \mathbf{\psi}, \mathbf{\gamma}, \mathbf{\lambda})
    \label{eq:elbo}
    \\
    &
    =
    \E_q[\log p(\mathbf{w}, \mathbf{z}, \mathbf{\theta}, \mathbf{\varphi}
    \mid \alpha, \beta)]
    -
    \E_q[\log q(\mathbf{z}, \mathbf{\theta}, \mathbf{\varphi})].
    \nonumber
  \end{align}
  Maximizing the ELBO is equivalent to minimizing the Kullback-Leibler
  divergence between $q(\mathbf{z}, \mathbf{\theta}, \mathbf{\varphi})$ and the
  posterior $p(\mathbf{z}, \mathbf{\theta}, \mathbf{\varphi} \mid
  \mathbf{w}, \alpha, \beta)$.

  ELBO $\LL$ can be optimized using coordinate ascent over the variational
  parameters (detailed derivation in \cite{Blei:2003, DeSmet:2009}):
  \begin{align}
    \psi_{m, v, k}
    &\propto
    \exp
    \left\{
      \E_q[\log\theta_{m, k}] + \E_q[\log \varphi_{k, v}]
    \right\}, \label{eq:psi}
    \\
    \gamma_{m, k}
    &
    =
    \alpha + \textstyle \sum_{v = 1}^V n_{m, v} \psi_{m, v, k},
    \label{eq:gamma}
    \\
    \lambda_{k, v}
    &
    =
    \beta + \textstyle \sum_{m = 1}^M  n_{m, v} \psi_{m, v, k},
    \label{eq:lambda}
  \end{align}
  where $n_{m, v}$ is the number of terms $v$ in document $m$. The
  expectations are
  \begin{align*}
    \E_q[\log \theta_{m, k}]
    &
    =
    \Psi(\gamma_{m, k})
    -
    \Psi
    \left(
      \textstyle
      \sum_{\widetilde{k} = 1}^K \gamma_{m, \widetilde{k}}
    \right),
    \\
    \E_q[\log \varphi_{k, v}]
    &
    =
    \Psi(\lambda_{k, v})
    -
    \Psi
    \left(
      \textstyle
      \sum_{\widetilde{v} = 1}^V \lambda_{k, \widetilde{v}}
    \right),
  \end{align*}
  where $\Psi$ denotes the digamma function (the first derivative of the
  logarithm of the gamma function).

  The updates of the variational parameters are guaranteed to converge to a
  stationary point of the ELBO. We can make some parallels with
  Expectation-Maximization (EM) algorithm \cite{Dempster:1977}. Iterative
  updates of
  $\mathbf{\gamma}$ and $\mathbf{\psi}$ until convergence, holding
  $\mathbf{\lambda}$ fixed, can be seen as ``E''-step, and updates of
  $\mathbf{\lambda}$, given $\mathbf{\gamma}$ and $\mathbf{\psi}$, can be seen
  as ``M''-step.

  The Variational Bayesian inference algorithm first initializes
  $\mathbf{\lambda}$ randomly. Then for each documents does the ``E''-step:
  initializes $\mathbf{\gamma}$ randomly and then until $\mathbf{\gamma}$
  converges does the coordinate ascend using Equations \eqref{eq:psi} and
  \eqref{eq:gamma}. After $\mathbf{\gamma}$ converges, the algorithm performs the
  ``M''-step: sets $\mathbf{\lambda}$ using Equation
  \eqref{eq:gamma}. Each combination ``E'' and ``M''-step improves
  ELBO. Variational Bayesian inference finishes after relative improvement of
  $\LL$ is less than a pre-prescribed limit or after we reach maximum number of
  iterations. We define an iteration as ``E'' + ``M''-step.

  After algorithm finishes $\mathbf{\gamma}$ represents documents topics and
  $\mathbf{\lambda}$ represents corpus topics.

  \SUBSEC{Online Variational Bayesian inference}

  Previously described algorithm has constant memory requirements. It requires
  full pass through the entire corpus each iteration. Therefore, it is not
  naturally suited when new data is constantly arriving. We would
  like an algorithm that gets the data, calculates the data topics and updates
  the existing corpus topics.

  Let us modify previous algorithm and make desired one. First, we factorize ELBO \pdfcomment{Kaj pa referenca? previous algorithm}
  (Equation \eqref{eq:elbo}) into:
  \begin{align*}
    \LL&(\mathbf{w}, \mathbf{\psi}, \mathbf{\gamma}, \mathbf{\lambda})
    =
    \\
    &\textstyle \sum_{m = 1}^M
    \left\{
      \E_q[ \log p(w_m \mid \theta_m, z_m, \mathbf{\varphi})]
      +
      \E_q[ \log p(z_m \mid \theta_m)]
    \right.
    \\
    &-
    \E_q[ \log q(z_m)]
    +
    \E_q[ \log p(\theta_m \mid \alpha)]
    -
    \E_q[ \log q(\theta_m)]
    \\
    &+
    \left.
    \left(
      \E_q[\log p(\mathbf{\varphi} \mid \beta)]
      -
      \E_q[\log q(\mathbf{\varphi})]
    \right)/M
    \right\}.
  \end{align*}
  Note that we bring the per corpus topics terms into the summation over
  documents, and divide them by the number of documents $M$. This allows us to
  look at the maximization of the ELBO according to the parameters
  $\mathbf{\psi}$ and $\mathbf{\gamma}$ for each document individually. Therefore, we first maximize ELBO according to the $\mathbf{\psi}$
  and $\mathbf{\gamma}$ as in previous algorithm with $\mathbf{\lambda}$
  fixed. Then we choose such
  $\mathbf{\lambda}$ for which the ELBO is as high as possible. Let
  $\gamma(w_m, \mathbf{\lambda})$ and
  $\psi(w_m, \mathbf{\lambda})$ be the values of $\gamma_m$
  and $\psi_m$ produced by the ``E''-step. Our goal is to find $\mathbf{\lambda}$
  that maximizes
  \begin{align*}
    \LL(\mathbf{w}, \mathbf{\lambda})
    =
    \textstyle
    \sum_{m = 1}^M
    \ell_m\left(w_m, \gamma(w_m, \mathbf{\lambda}),
    \psi(w_m, \mathbf{\lambda}), \mathbf{\lambda}\right),
  \end{align*}
  where
  $
  \ell_m(w_m, \gamma(w_m, \mathbf{\lambda}),
  \psi(w_m, \mathbf{\lambda}), \mathbf{\lambda})
  $
  is the $m$-th document's contribution to ELBO.

  Then we compute $\mathbf{\widetilde{\lambda}}$, the setting of $\mathbf{\lambda}$
  that would be optimal with given $\psi$ if our entire corpus consisted of the
  single document $w_m$ repeated $M$ times:
  \begin{align*}
    \widetilde{\lambda}_{k, v}
    =
    \beta + Mn_{m, v}\psi_{m, v, k}.
  \end{align*}
  Here $M$ is the number of available documents, the size of the corpus. Then we
  update $\mathbf{\lambda}$ using convex combination of its previous
  value and $\mathbf{\widetilde{\lambda}}$:
  $\mathbf{\lambda}
  =
  (1 - \rho_m)\mathbf{\lambda} + \rho_m \mathbf{\widetilde{\lambda}}$, where the
  weight is $\rho_m = (\tau_0 + m)^{-\kappa}$. Unknowns have special meaning:
  $\tau_0 \geq 0$ slows down the early iteration and $\kappa$ controls the rate
  at which old values $\mathbf{\widetilde{\lambda}}$ are forgotten.

  To sum up. The algorithm firstly initializes $\mathbf{\lambda}$ randomly. Then, on a
  given document, performs ``E''-step as in Variational Bayesian
  inference. Next it updates $\mathbf{\lambda}$ as discussed above.
  Finally it moves on the new document and repeats everything. The algorithm
  terminates after all documents are processed.

  This algoritem is called Online Variational Bayesian inference (Online VB) and
  was proposed by Hofffman, Blei and Bach in \cite{Hoffman:2011}.

  \SEC{EXPERIMENTS}

  We ran several experimets to evaluate algorithms of the LDA model. Our purpose
  was to compare the time complexity and performance of previously described
  algo\-ri\-thms. \pdfcomment{The implementation is as similar as possible.} For training and
  testing corpora we used Wikipedia.

  Efficiency was measured by using perplexity on held-out data, which is defined
  as
  \begin{align*}
    \mbox{perplexity}(\mathbf{w}_{\mbox{test}}, \mathbf{\lambda})
    =
    \exp
    \left\{-
      \frac{\sum_{m = 1}^M \log p(\mathbf{w}_m \mid \mathbf{\lambda})}
      {\sum_{m = 1}^M N_m}
    \right\},
  \end{align*}
  where $N_m$ denotes number of words in $m$-th document. Since we cannot
  directly compute $\log p(\mathbf{w}_m \mid \mathbf{\lambda})$, we use ELBO as
  approximation:
  \begin{align*}
    &
    \mbox{perplexity}(\mathbf{w}_{\mbox{test}}, \mathbf{\lambda})
    \\
    &
    \leq
    \exp
    \left\{
      -
      \textstyle \sum_{m = 1}^M
      (\E_q[\log p(\mathbf{w}_m, \mathbf{z}_m, \mathbf{\theta}_m
      \mid \mathbf{\varphi})]
    \right.
    \\
    &
    \quad-
    \left.
    E_q[\log q(\mathbf{z}_m, \mathbf{\theta}_m \mid \mathbf{\varphi})])
    \right / \textstyle \sum_{m = 1}^M N_m \}.
  \end{align*}
  We tested three algorithms and ran experiments for 10.000,\\ 20.000, \ldots, 80.000
  documents as a training set for corpora topics. Later we evaluated
  perplexity on 100 held-out documents. Size of vocabulary was around 150.000
  words. 

  In all experiments $\alpha$ and $\beta$ are fixed at $0.01$ and the number of
  topics $K$ is equal to $100$. For Collapsed Gibbs sampling, no experiment
  converged. The criteria was relative change in $\mathbf{z}$ variable; change did
  not get under 20\% in 1000 iterations.

In Variational Bayesian inference the ``E''-step and  the ``M''-step converge  if relative change in $\gamma$ is under $0.001$ and 
relative improvement of the ELBO is under $0.001,$ respectively.
If there is no convergence, we terminate after $100$ iterations
  for both ``E'' and ``M''-step. However algorithm always converged in less than 20 iterations.

  In Online Variational Bayesian inference limit for the ``E''-step was the
  same as in Variational Bayesian inference. Batchsize was 100 documents,
  $\tau_0$ was 1024 and $\kappa$ was equal to $0.7$ as proposed in \cite{Hoffman:2011}.

  \vspace{-3mm}

  \begin{figure}[!ht]
    \centering
    \includegraphics[scale=0.42]{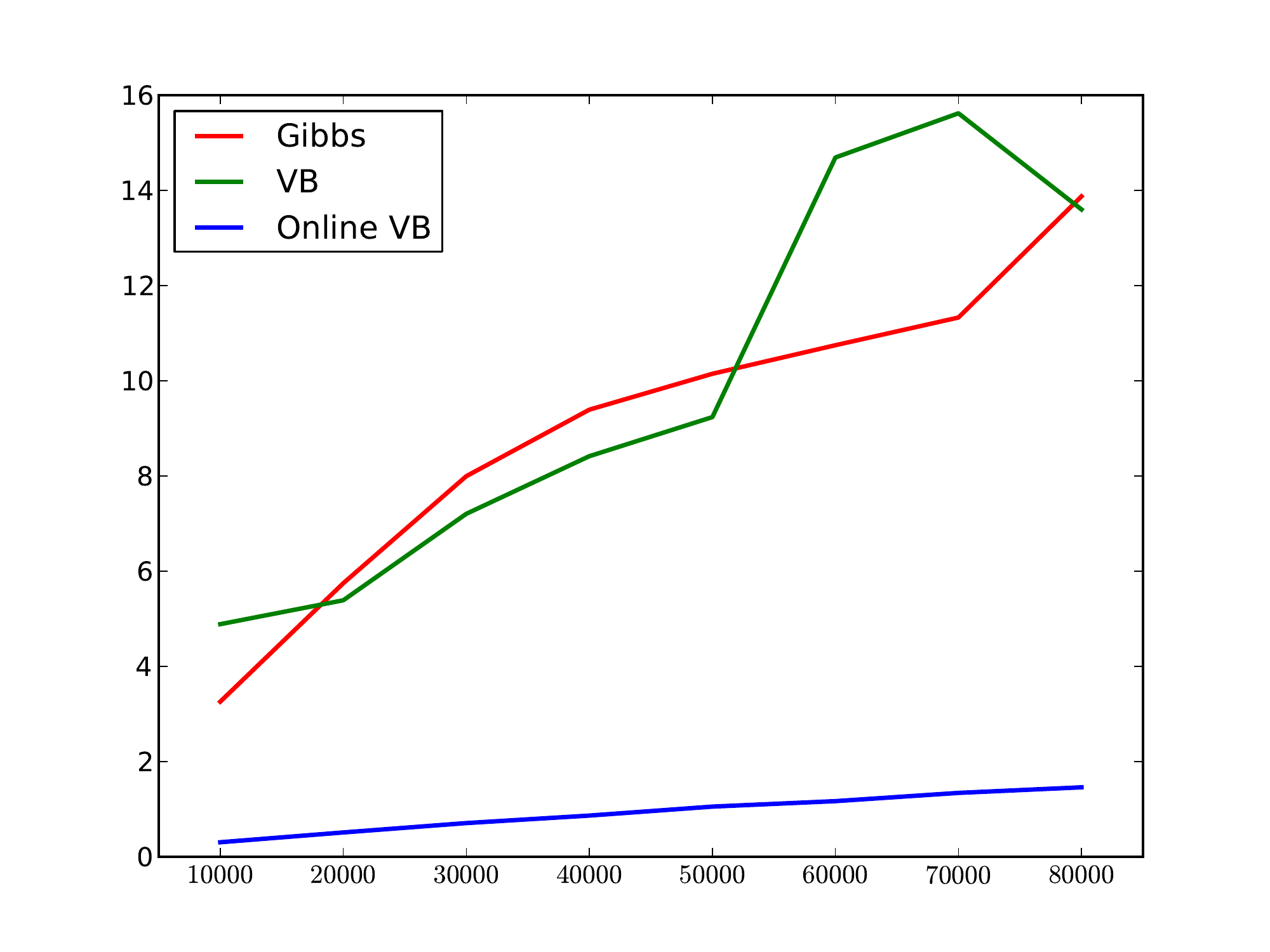}
    \vspace{-10mm}
    \caption{Time used by the algorithms (in hours) given the number of the documents.}
    \label{fig:time}
  \end{figure}

  \begin{figure}[!ht]
    \centering
    \includegraphics[scale=0.42]{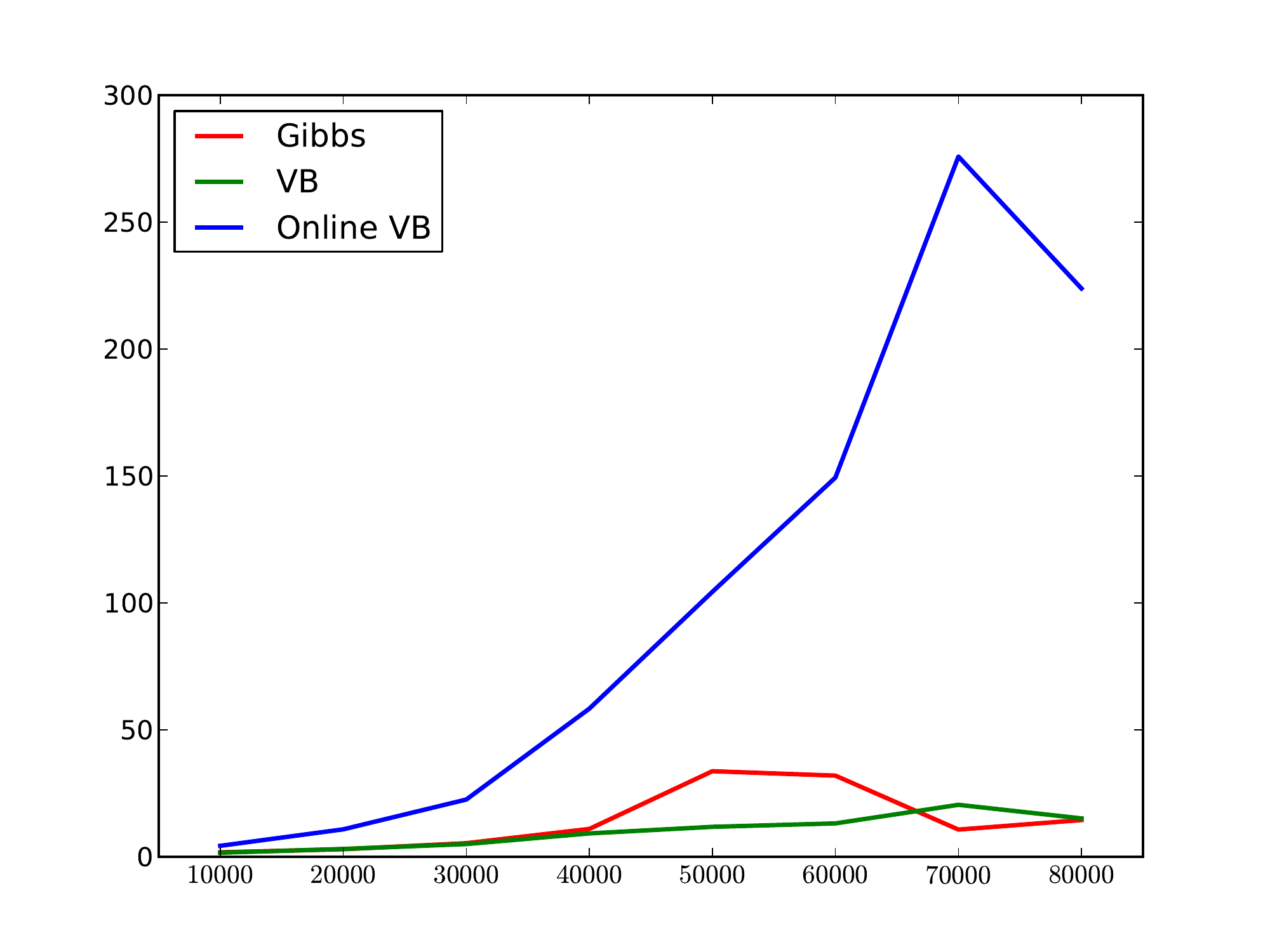}
    \vspace{-10mm}
    \caption{Perplexity on held-out documents as a function of  number of
      documents analyzed.}
    \label{fig:perplex}
  \end{figure}

  The fastest algorithm is Online VB, other two have similar time complexity
  with a large note: VB algorithm converged every time while Gibbs
  sampling algorithm did not converge. Unexpected, Online VB does not perform as
  well as other two but in practice still gives reasonably good\pdfcomment{Explain unexpected}
  results. Our future goal is to explain the results obtained by experiments.
  
  \pdfcomment{We see that Gibbs sampling and VB are similar with respect to
  performance. But if Gibbs sampling converged, we would expect its better
  performance.}

  Therefore we recommend Online VB algorithm for practical use, if time
  is a factor.


  \vspace{-4mm}

  \SEC{ACKNOWLEDGMENT}

  The authors gratefully acknowledge that the funding for this work was provided
  by the project XLIKE (ICT-\\257790-STREP).

  \vspace{-5mm}

  \bibliography{lda_algorithms}{}
  \bibliographystyle{plain}


\end{document}